\documentclass[twoside,11pt]{article}

\usepackage{jmlr2e}
\usepackage{natbib}
\usepackage{amsmath}

\ShortHeadings{Kernel Alignment to Handle Mixed Data}{Handhayani and Cussens}
\firstpageno{1}

\begin{document}

\title{Kernel-based Approach to Handle Mixed Data for Inferring Causal Graphs}

\author{\name Teny Handhayani \email th1075@york.ac.uk \\
       \addr Department of Computer Science\\
       University of York\\
       Deramore Lane, Heslington York  YO10 5GH, United Kingdom
       \AND
       \name James Cussens \email james.cussens@york.ac.uk \\
       \addr  Department of Computer Science\\
       University of York\\
       Deramore Lane, Heslington York  YO10 5GH, United Kingdom}

\editor{}

\maketitle

\begin{abstract}
	
Causal learning is a beneficial approach to analyze the cause and effect relationships among variables in a dataset. A causal graph can be generated from a dataset using a particular causal algorithm, for instance, the PC algorithm or Fast Causal Inference (FCI). Generating a causal graph from a dataset that contains different data types (mixed data) is not trivial. This research offers an easy way to handle the mixed data so that it can be used to learn causal graphs using the existing application of the PC algorithm and FCI. This research proposes using kernel functions and Kernel Alignment to handle a mixed data. Two main steps of this approach are computing a kernel matrix for each variable and calculating a pseudo-correlation matrix using Kernel Alignment. Kernel Alignment is used as a substitute for the correlation matrix for the conditional independence test for Gaussian data in PC Algorithm and FCI. The advantage of this idea is that is possible to handle any data type by using a suitable kernel function to compute a kernel matrix for an observed variable. The proposed method is successfully applied to learn a causal graph from a mixed data containing categorical, binary, ordinal, and continuous variables. 

\end{abstract}

\begin{keywords}
  Causal learning, Kernel function, Kernel Alignment, PC algorithm, FCI
\end{keywords}

\section{Introduction}

Causal analysis is inferring probabilities through changing condition, for example, treatment or external intervention~\citep{Pearl10}. A distribution function does not show how the distribution will be different if external conditions are changed. It happens because the laws of probability theory do not instruct how one variable should change when another variable is modified \citep{Pearl10}. This problem can be solved by causal assumptions which identify relationships that remain the same when external conditions change. Causal learning and analysis are beneficial to answer the cause-effect questions in some areas, for instance, medical treatment, health, economic, and politics. The studies in inferring causal graph in health and medical treatments have been done in the previous research by \citet{Hill16}, \citet {Kaiser16}, \citet{Peng14}, \citet{Triantafillou17}, and \citet{Yang14}. Causal inference in economics can be read in some publications by \citet{Brodersen15}, \citet{Varian16}, and \citet{Zhang14}. Causal inference in political science has been studied by \citet{Blackwell13}.

A causal graph is a graphical model used to describe the cause-effect relationship among variables or events. Causal graphs can be generated from datasets using causal algorithms.  There are some popular causal algorithms, for example, Greedy Equivalence Search \citep{Chickering02}, Max-Min Hill-Climbing \citep{Tsamardinos06}, PC algorithm \citep{CausationPS}, Fast Causal Inference (FCI)  \citep{CausationPS} and Really Fast Causal Inference (RFCI) \citep{Colombo12}. 

A learned graph is an outcome of causal learning inference from the dataset using a particular algorithm. In fact, real datasets do not always have homogenous data types. Real datasets are the datasets collected from real events, for example, patient records. Patient record datasets might comprise various variables with different data types: gender (binary), height (continuous), weight (continuous), blood pressure (continuous), and blood type (categorical). A dataset that consists of variables which have different data types is called heterogeneous data or mixed data. On the other hand, a dataset that has the same data type for all variables (continuous or discrete) is named homogeneous data. PC algorithm and FCI need conditional independence test to generate a causal graph from the dataset \citep{CausationPS,Kalisch07}. Conditional independence test is easy if the variables have the same data type. However, if the variables have different data types, conditional independence test is not trivial. 

There are some useful applications for learning causal graphs from datasets, for instance, \textit{pcalg}. \textit{pcalg} is an applications for graphical model and causal learning \citep{Kalisch12}. It has the implementation of popular causal algorithms, for instance, PC algorithm, FCI, and RFCI \citep{Kalisch12}. PC algorithm and FCI can be used to learn causal graphs as long as it exists a method for conditional independence test suitable to the datasets.  

Copula PC algorithm is one of a variant of the PC algorithm and it can be used to learn causal graphs from mixed data containing binary, ordinal, and continuous variables \citep{Cui16}. The Copula PC algorithm is a method for causal discovery from mixed data with an assumption the data is drawn from a Gaussian copula model \citep{Cui16}. It implements the Gibbs sampling procedure based on the extended rank likelihood, then estimates the scale matrix and the degree of freedom from the Gibbs samples. The scale matrix substitutes for a correlation matrix and the degree of freedom acts as an effective number of data points for the conditional independence test \citep{Cui16}. If it is used to learn causal graphs from mixed data where some of the variables are categorical variables, the rank procedure might reveal a problem.

This research is conducted to solve the problem in learning causal graphs from mixed data using PC algorithm/FCI and offers the proper treat for categorical variables as well as binary, ordinal, and continuous variables. The research question is how to treat mixed data containing categorical, binary, ordinal, and continuous variables when inferring a causal graph? 

We propose kernel function and Kernel Alignment to handle mixed data. The goal of this research is to handle the mixed data so that it can be used to learn the causal graphs using the existing application. This research offers a simple way to learn a causal graph from a mixed data using the existing application for PC algorithm and FCI from \textit{pcalg}. In this research, we use the PC algorithm and FCI application from \textit{pcalg} \citep{Kalisch12}. The novelty of this research is using kernel functions and Kernel Alignment to handle mixed data. This proposed method is useful for learning a causal graph from mixed data.

\section{Literature Review}

\subsection{PC algorithm}
\begin{figure}[ht]
	\centering
	\includegraphics[width= 8 cm]{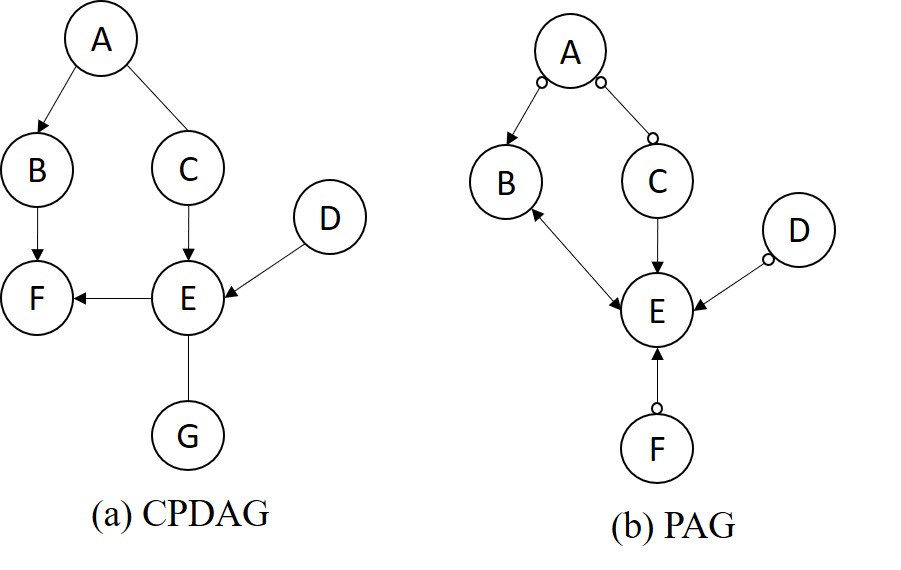}
	\caption{The example of a CPDAG and a PAG}
	\label{fig:ExmCPDAGPAG}
\end{figure}

PC algorithm is a causal inference algorithm introduced by Peter Spirtes and Clark Glymour \citep{CausationPS}. Generally, PC algorithm consists of two main stages: generating skeleton and orienting the edges. The detailed procedures of the PC algorithm has been explained by \citet{Colombo14}. The PC algorithm identifies the equivalence class of Bayesian Network in polynomial time if the graph structure is a Directed Acyclic Graph (DAG) and each node has a limited degree \citep{Chickering15}. The output of the PC algorithm is represented by Completed Partially Directed Acyclic Graph (CPDAG) \citep{Kalisch07}. Figure \ref{fig:ExmCPDAGPAG} (a) shows an example of a graph represented by a CPDAG. PC algorithm is a causal algorithm that can be applied to a dataset assuming there are no latent variables. A latent variable is a variable that is not measured or not recorded \citep{Colombo12}.

PC algorithm applies a conditional independence test to generate a graph from a dataset. A conditional independence test for Gaussian data can be done using equation \ref{condtest} \citep{Cui16}. It tests ‘is a variable $Z_{u}$ independent $Z_{v}$ given $Z_{S}$?, where $n$ is the number of samples, $S$ is the separation set, $\hat{\rho}$ is partial correlation, $\alpha$ is a significance level, and $\Phi$ is Gaussian cumulative distribution function (cdf) \citep{Cui16}. Assume, the distribution \textit{P} of the random variables \textit {X} is multivariate normal. For $i \neq j \in \{1,...,p\}, k \subseteq \{1,...,p\} \backslash \{i,j\} $ and the partial correlation between $X_{i}$ and $X_{j}$ given $\{X_{r};r \in k\}$ is denoted by $\rho_{i,j|k}$. The $ \rho_{i,j|k}= 0$ if and only if $X_{i}$ and $X_{j}$ are conditionally independent given $\{X_{r};r \in k\}$. The estimated partial correlation $\hat{\rho}_{i,j|k} $ can be computed via the regression, inversion of part of the covariance matrix or recursively. Fisher’s z-transform is used to test whether the partial correlation is equal to zero or not. We reject the null hypothesis $H_{0}(i,j|k):\rho_{i,j|k} = 0$ against the two-sided alternative $H_{A}(i,j|k):\rho_{i,j|k} \neq 0$ if $\sqrt{n-|k|-3}|Z(i,j|k)|> \Phi^{-1}(1-\alpha/2)$ \cite{Kalisch07}.

\begin{equation}
\label{condtest}
Z_{u}\bot Z_{v}|Z_{S} \Leftrightarrow \sqrt{n-|S|-3} \left| \frac{1}{2} \log \left( \frac{1+\hat{\rho}_{uv|S}}{1-\hat{\rho}_{uv|S}} \right) \right| \leq \Phi ^{-1}(1-\alpha/2).
\end{equation}

\subsection{Fast Causal Inference}
Fast Causal Inference (FCI) is the improvement algorithm from PC algorithm \citep{CausationPS}. The early steps of FCI is generating a skeleton graph then orienting the edges. \citet{Zhang082} has explained the detail orientation rule for FCI. The advantage of FCI is that it can detect the latent variables. \citet{Colombo12} elucidated the detail procedure of FCI. FCI produces a causal graph that is represented as a Partial Ancestral Graph (PAG) \citep{Colombo12}. A PAG has three different marks to form six type of edges: $\circ\!\!\!\rightarrow$, $\leftrightarrow$, $\circ\!\!-\!\!\circ$, $\rightarrow$, $\circ\!-$, and $\!\!-\!\!-$ \citep{Zhang082}. Figure \ref{fig:ExmCPDAGPAG} (b) shows an example of a PAG. FCI is a causal algorithm that is able to generate a causal graph from the dataset that might contain latent variables.

\subsection{Structural Hamming Distance}

Structural Hamming Distance (SHD) is a method to evaluate the structure of a learned graph. SHD computes the mismatch of the structure a learned graph to the true graph. The SHD score $S$ is a summation of extra edges, missing edges, and wrong orientation edges \citep{Tsamardinos06}.  A true edge is an edge in learned graph that exists in the true graph. A missing edge is an edge that exist in the true graph but it disappears in learned graph. A wrong mark edge is a true edge in learned graph but it has different mark in one or both of its sides. An extra edge is an edge that does not exist in true graph but it appears in learned graph. The normalized SHD score $S_{norm}$ can be computed using $S_{norm} = \frac{S}{\binom{n}{2}/2}$, where $\binom{n}{2}$ is binomial coefficient and \textit{n} is the number of observed variables in a dataset \citep{Malone15}. The lower SHD score means that the learned graph has less mismatch to the true graph. It means that the learned graph has a high similarity to the true graph.

\section{Kernel Function and Kernel Alignment to Handle Mixed Data}

PC algorithm and FCI in \textit{pcalg} need a correlation matrix as an input for conditional independence test \citep{Kalisch12}. Computing correlation matrix from two variables with different data types (i.e, continuous and categorical) is impossible. We propose an easy way to compute a pseudo correlation matrix from the mixed data through a kernel-based approach. The idea of the proposed method is transforming mixed data into homogeneous data, then compute a correlation matrix. The detail of the proposed method is explained as follows. First, each variable from the mixed data is treated individually by computing its kernel matrix using a suitable kernel function for its data type. The kernel matrices represent the transformed data of variables. The kernel function is applied as a method for transforming the heterogeneous dataset into the homogeneous dataset because kernel matrices have a numeric data type. The transformed data are set of kernel matrices, so computing a correlation matrix cannot be done using a common method, for example, Pearson Correlation. 

A correlation coefficient measures the strength of relationships between two variables \textit{X} and \textit{Y} by a single number and it has the range value of [-1 ,1] \citep{ProbabilityStat}. Suppose \textit{X} and \textit{Y} are variables in input space and their relationship can be measured by correlation coefficient. Suppose, a set of kernel matrices is the new form of mixed data in the feature space. When \textit{X} and \textit{Y} are mapped into the feature space, their relationship can be measured by the Kernel Alignment. Therefore, the authors propose Kernel Alignment to compute the pseudo-correlation matrix from the set of kernel matrices.

\subsection{Kernel Function}

A kernel is a function $\kappa$ that for all  $x,z \in X$ satisfies $\kappa(x,z)=\left \langle \phi (x),\phi(z) \right\rangle$, where $\phi$ is a mapping from \textit {X} to an (inner product) feature space \textit {F}, $\phi:x\rightarrow \phi(x) \in F$. Suppose a set of vectors $S= \{x_{1},...,x_{\ell} \}$, the kernel matrix is $\ell \times \ell$ matrix \textit{G} whose entries are $G_{ij} =  \langle x_{i},x_{j} \rangle$. A kernel matrix is positive semidefinite \citep{KernelM}.

This research implements RBF kernel and Categorical kernel. RBF kernel is a polynomial kernel of infinite degree \citep{KernelM}. RBF Kernel and Categorical kernel can be computed using equation \ref{rbfkern} \citep{KernelM} and equation \ref{catkern} \citep{Belanche13}, respectively. \\ \\
RBF Kernel:
\begin{equation}
\label{rbfkern}
\kappa \left(x_{i},x_{j} \right) = \exp \left( -\frac{\left \Arrowvert x_{i}-x_{j} \right \Arrowvert^2}{2\sigma^2} \right), \sigma > 0
\end{equation}
Categorical Kernel:
\begin{equation}
\label{catkern}
\kappa \left( z_{i},z_{j}\right) = \left\{ \begin{array}{ll} 
h_{\theta}\left(P_{Z} \left(z_{i}\right) \right) & \mbox{if $z_{i}=z_{j}$}\\
0 & \mbox{if $z_{i}\neq z_{j}$}.\end{array} \right.
\end{equation}

P is probability and $h_{\theta}(.)$ is a function that depends on the parameter $\theta$. It is defined as  $h_{\theta}(z) = \left( 1-z^\theta \right)^{1/\theta}, \theta >0$ \citep{Belanche13}. 

The kernel matrix consist of the scalar product of the data points in feature space. In the feature space, the matrix K is called in ‘ill-condition’ if it has about the same value because the origin is far away from the convex hull of the data \citep{Meila03}. To solve this problem, it needs centering data.  Centering the data in the feature space can be done by centering the kernel matrix \citep{KernelM}. The goal of centering data is transferring the origin of the feature space to the center of mass of the training examples. This research implements a method for centering kernel matrix developed by Shawe-Taylor and Cristianini \citep{KernelM}.

\subsection{Kernel Alignment}

Originally, Kernel Alignment was applied to measure the similarity between two kernel functions or between a kernel and a target function \citep{Cristianini01}. Suppose, it is given a sample $S = \{x_{1},...,x_{m} \}$, the inner product between two kernel matrices is $\left \langle K_{1},K_{2} \right\rangle = \sum_{i,j=1}^{m} K_{1}(x_{i},x_{j}) K_{2}(x_{i},x_{j})$. The alignment between a kernel $k_{1}$ and $k_{2}$ of the sample \textit{S} is defined as equation \ref{kaeq} where $K_{i}$ is the kernel matrix of the sample \textit{S} using kernel function  $\kappa_{i}$ \citep{Cristianini01}. 

\begin{equation}
\label{kaeq}
\hat{A}(S,k_{1},k_{2})=\frac{\langle K_{1},K_{2} \rangle}{\sqrt{\langle K_{1},K_{1} \rangle \langle K_{2},K_{2} \rangle}}
\end{equation} 


The alignment satisfies $-1\leq A(K_{1},K_{2})\leq 1$ because it can be viewed as the cosine of the angle between $\ell^2$-dimensional vectors of $\ell \times \ell$ matrices \citep{KernelM}. If $K_{1}$ and $K_{2}$ are positive semi definite, the value of $\langle K_{1},K_{2} \rangle \geq 0$ \citep{KernelM}. In consequence Kernel Alignment has range of values $0\leq A(K_{1},K_{2})\leq 1$. 

\begin{figure}[ht]
	\centering
	\includegraphics[width=\textwidth]{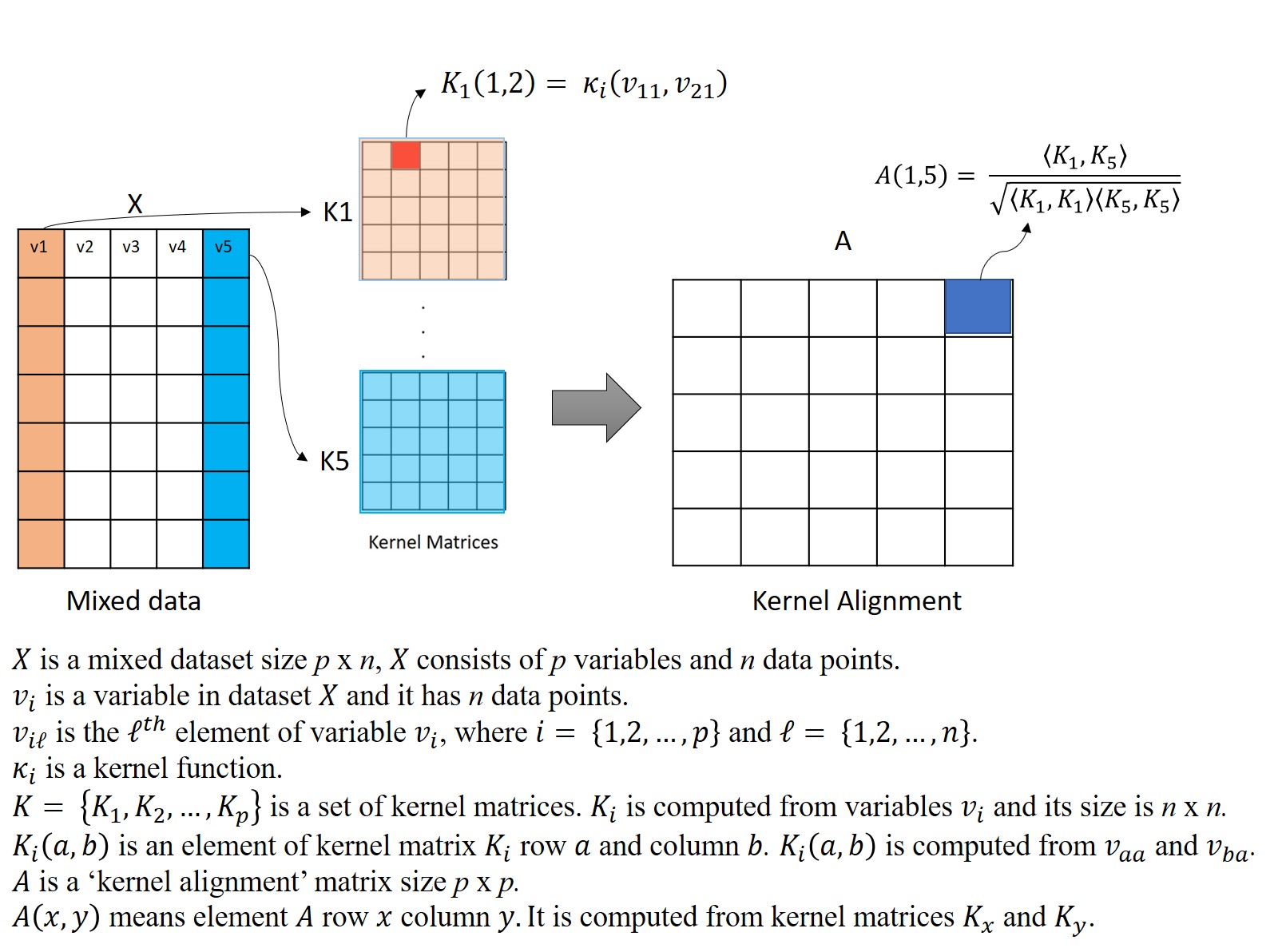}
	\caption{Illustration of the kernel alignment approach}
	\label{fig:KAmethod}
\end{figure}

Figure \ref{fig:KAmethod} shows the illustration of the kernel alignment approach. Suppose, the mixed dataset \textit{X} consist of \textit{p} variables and \textit{n} data points. For $i \in \{1,...,p\}$, we compute a kernel matrix $K_{i}$ for each variable $X_{i}$ using a suitable kernel function, $K_{i}(j,l)=K_{i} (x_{ji},x_{li})$.  The alignment $A(i,j)$ is computed from two kernel matrices $K_{i}$ and $K_{j}$. Equation \ref{kadet} shows the detailed implementation of Kernel Alignment, where, $K_{z=i}$ is the kernel matrix \textit{i}, $K_{z=i}(a,b)$ is an element of kernel matrix row \textit{a} and column \textit{b}, and $\kappa_{z=i}$ is a kernel function \textit{i}. The kernel matrices represent variables in space and the alignment measures their relationship. Kernel Alignment produces a pseudo-correlation matrix. It is treated just like a normal correlation matrix.

\begin{equation}
\label{kadet}
\renewcommand{\arraystretch}{1.3}
\begin{array}{l@{\;}l}
A(i,j)= \frac{ \langle K_{z=i},K_{z=j} \rangle}{\sqrt{\langle K_{z=i},K_{z=i} \rangle \langle K_{z=j},K_{z=j} \rangle}}\\
\\
A(i,j)= \frac{\sum_{a,b = 1}^{n}K_{z=i}(a,b)K_{z=j}(a,b)}{\sqrt{\left(\sum_{a,b = 1}^{n}K_{z=i}(a,b)K_{z=i}(a,b)\right)\left(\sum_{a,b = 1}^{n}K_{z=j}(a,b)K_{z=j}(a,b)\right)}}\\
\\
A(i,j)= \frac{\sum_{a,b=1}^{n}\kappa_{z=i}\left(x_{ai},x_{bi}\right)\kappa_{z=j}\left(x_{aj},x_{bj}\right)}{\sqrt{\left(\sum_{a,b=1}^{n}\kappa_{z=i}\left(x_{ai},x_{bi}\right)\kappa_{z=i}\left(x_{ai},x_{bi}\right)\right)\left(\sum_{a,b=1}^{n}\kappa_{z=j}\left(x_{aj},x_{bj}\right)\kappa_{z=j}\left(x_{aj},x_{bj}\right)\right)}}
\end{array}
\end{equation}

The idea of implementing the Kernal Alignment as the pseudo-correlation matrix is similar to a concept of measuring dependence between random variables using distance correlation. Therefore, measuring dependencies using Kernel Alignment is similar to measuring dependencies using distance correlation. Distance correlation satisfies $0 \leq \mathcal{R} \leq 1$ and $\mathcal{R} = 0$ if \textit{X} and \textit{Y} are independent \citep{Szekely07}. Distance correlation $\mathcal{R}\left(X,Y\right)$ can be implemented if \textit{X} and \textit{Y} have same data types. On the other hand, Kernel Alignment can be applied for variables with different data types. Kernel Alignment can be viewed as a distance correlation and it is used for conditional independence test.  

Conditional independence test for the proposed method follows the rule in equation \ref{condtest}. Formally, partial correlation between two variables $X_i$ and $X_j$ given $Y= ( Y_1,Y_2,…,Y_k )$ is the correlation between residual $e_{(X_i)}$ resulting from the linear regression of $X_i$ with $Y$ and $e_{(X_j)}$ producing from the linear regression of $X_j$ with $Y$. Partial correlation has range value [-1,1]. The partial correlation $\hat{\rho}$ can be computed from the correlation matrix using equation \ref{parcor} \citep{Kalisch07}. Suppose, $A$ is the Kernel Alignment and $C$ is the correlation matrix, Kernel Alignment $A$ has a range value [0, 1] thus it can be said that $A \subseteq C$. When the Kernel Alignment substitutes the correlation matrix then the partial correlation is computed from Kernel Alignment. Since $A \subseteq C$, the value of the partial correlation is still in boundaries [-1,1]. It shows that Kernel Alignment is acceptable to be used as a pseudo correlation matrix and participate in a conditional independence test. 

\begin{equation}
\label{parcor}
\hat{\rho}_{i,j|k}= \frac{\rho_{i,j|k\backslash h}-\rho_{i,h|k\backslash h}\rho_{j,h|k \backslash h}}{\sqrt{\left(1-\rho_{i,h|k\backslash h}^{2} \right)  \left(1-\rho_{j,h|k\backslash h}^{2} \right)}} 
\end{equation}

\begin{figure}[ht]
	\centering
	\includegraphics[width=\textwidth]{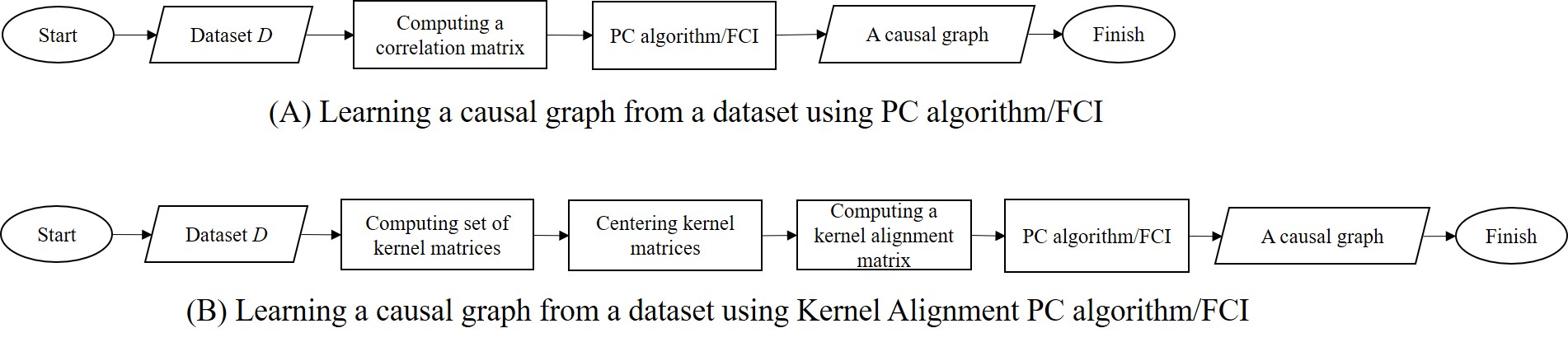}
	\caption{Comparison method to generate a causal graph using original PC algorithm/FCI and Kernel Alignment PC Algorithm (KAPC) or Kernel Alignment FCI (KAFCI)}
	\label{fig:OriPCvsKA}
\end{figure}

Kernel Alignment is implemented to replace the correlation matrix and it is used as an input for conditional independence test in PC algorithm and FCI. Figure \ref{fig:OriPCvsKA} shows a comparison way to generate a causal graph using original PC algorithm/FCI and Kernel Alignment PC algorithm/FCI in \textit{pcalg}. Figure \ref{fig:OriPCvsKA} shows two flowcharts that explain the original PC algorithm/FCI and KAPC/KAFCI. 

\section{Experiments}

\subsection{Datasets}
The mixed datasets are produced using forward sampling for a hybrid network. It implements Conditional Linear Gaussian (CLC) \citep{Pgm}. First, we generate some random DAGs consist of 10, 20 and 30 nodes. Each variable is given a data type. There are four different data types: categorical, binary, ordinal, and continuous. Then, the mixed data are generated based on these graphs.  For each different group of nodes, it is produced 10 different graphs and each graph obtained five different datasets. 

We generate two groups datasets according to the kind of data types. The first group is datasets consist of three mixed variables: discrete (binary and ordinal) and continuous variables. The second group is mixed dataset consists of four mixed variables: discrete (binary, ordinal, categorical) and continuous variables. The distribution of data type is around 15\% - 60\%. The generated datasets are used in the simulation. 

\subsection{Experimental Design}

The experiment is divided into several groups according to the algorithms and datasets. Table \ref{tb:ExpDesign} shows 3 group experiments. Experiment 1 uses mixed datasets containing binary, ordinal, and continuous variables. In the experiments, the pseudo correlation matrix is computed using the kernel-based approach, hereafter it is used as an input for a conditional independence test for the PC algorithm and FCI.  The purpose of this experiments is comparing the performance of Kernel Alignment PC algorithm (KAPC) to Copula PC algorithm. According to the previous research by \citet{Cui16}, Copula PC algorithm can be used to learn causal graph from mixed data containing binary, ordinal, and continuous variables. For fairness comparison, we do not use categorical variables. Experiment 2 and experiment 3 use mixed datasets containing categorical, binary, ordinal, and continuous variables. We only run KAPC and KAFCI for experiment 2 and experiment 3, respectively. The experiments are run in $\alpha = 0.1$ for conditional independence test. 

The detail implementation for learning a causal graph from a mixed data is explained as follow. The first step is computing kernel matrix from each variables using a suitable kernel function for each data type.  RBF kernel and Categorical kernel are applied to compute kernel matrices from continuous and discrete variables, respectively. Table \ref{tb:widthkern} shows the values of kernel parameters and they are chosen arbitrary.  The second step is computing a matrix Kernel Alignment from a set of kernel matrices. The third step is applying the Kernel Alignment matrix as input for conditional independence test to learn a causal graph using the PC algorithm/FCI. The quality of the structure of the learned graph is measured using normalized SHD score. 

The experiments use group of graphs containing 10, 20, and 30 variables. Each group of graphs has 10 different graphs and each graph has 5 different datasets, so overall there are 150 datasets. Some variables are deleted from the datasets for experiments using FCI. The deleted variables act as the latent variables. We randomly delete 1, 4, and 8 variables from graph 10, 20, and 30 nodes, respectively. There is no missing value in the dataset. The experiments are run for 100, 500, 1000, 1500, 2000, and 5000 data points. 
 
\begin{table}[ht]
	\caption{Experimental Design}
	\begin{tabular}{|l|l|l|l|l|}
		\hline
		Experiments  &  Kernel Alignment   & Copula     & Dataset         \\
		\hline
		Experiment 1 &  KAPC               & Copula PC  & Dataset Group 1 \\
		Experiment 2 &  KAPC               & -          & Dataset Group 2 \\
		Experiment 3 &  KAFCI              & -          & Dataset Group 2 \\
		\hline
	\end{tabular}
	\label{tb:ExpDesign}
\end{table}

\begin{table}[ht]
	\caption{The value of width RBF kernel $\sigma$ and  categorical kernel $\theta$.}
	\begin{tabular}{|l|l|l|l|l|l|l|l|l|l|}
		\hline
		Kernel Parameter            &   P1    & P2   & P3  & P4    & P5   & P6  & P7    & P8   & P9\\
		\hline
		RBF kernel $\sigma$         &  0.001  & 0.01 & 0.1 & 0.001 & 0.01 & 0.1 & 0.001 & 0.01 & 0.1\\
		Categorical kernel $\theta$ & 0.5     & 1    & 1.5 & 1     & 1.5  & 0.5 & 1.5   & 0.5  &  1\\
		\hline
	\end{tabular}
	\label{tb:widthkern}
\end{table}

\subsection{Experimental Result using Mixed Data: Binary, Ordinal, and Continuous Variables}

This section shows experiment results of Kernel Alignment PC algoritm (KAPC) using mixed datasets containing binary, ordinal and continuous variables. The goal of this experiment is observing the performance of kernel-based approach to handle mixed data for learning causal graphs. The experiment results of a kernel-based approach are compared to the result of the Copula PC algorithm \citep{Cui16}. We run the application of Copula PC algorithm from \citet{Cui16}. 

\begin{figure}[ht]
	\centering
	\includegraphics[width=\textwidth]{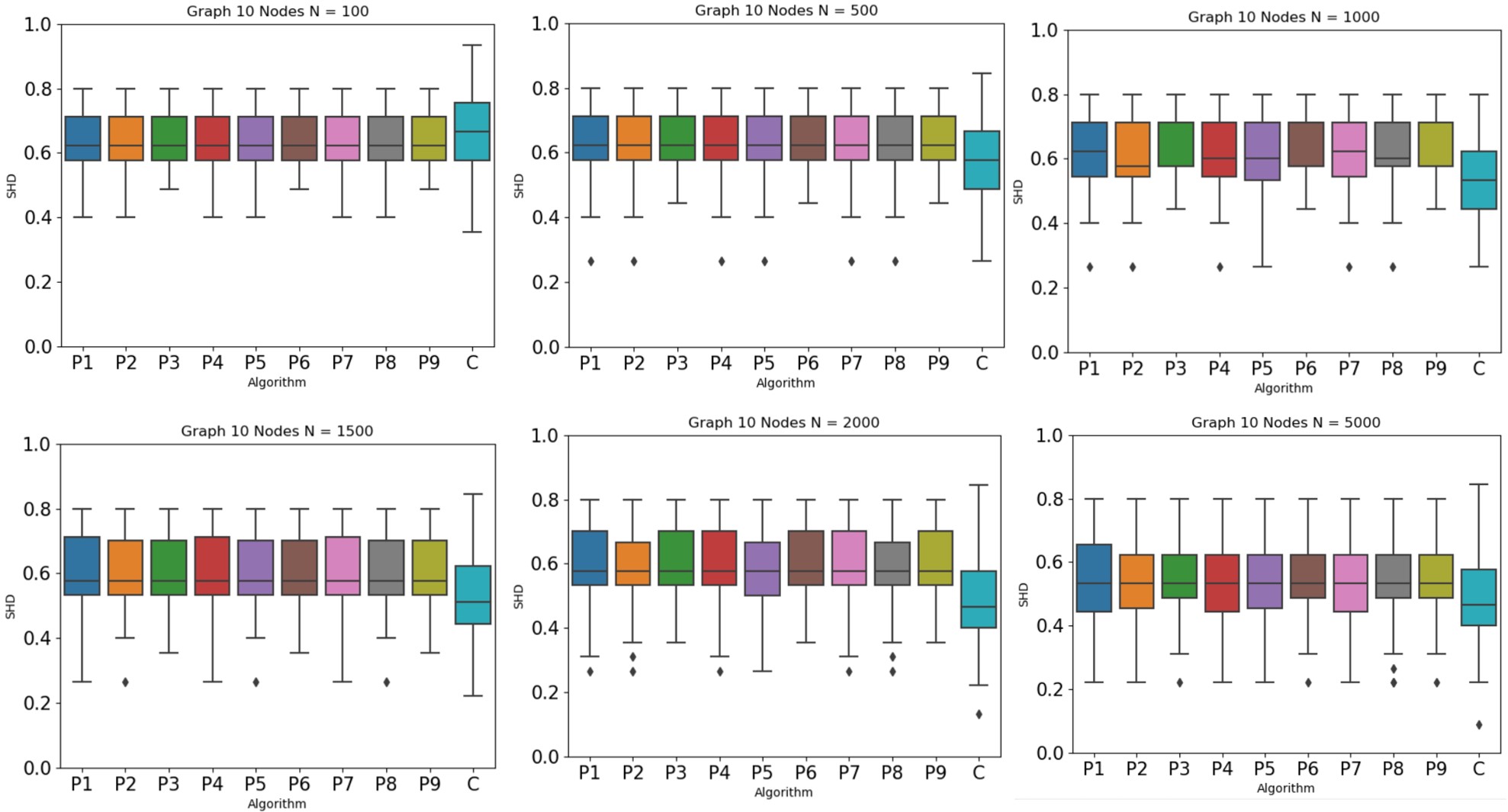}
	\caption{KAPC (P1-P9) and Copula PC (C) Graph Nodes = 10}
	\label{fig:KACopulaPC10_Exp1}
\end{figure}

\begin{figure}[ht]
	\centering
	\includegraphics[width=\textwidth]{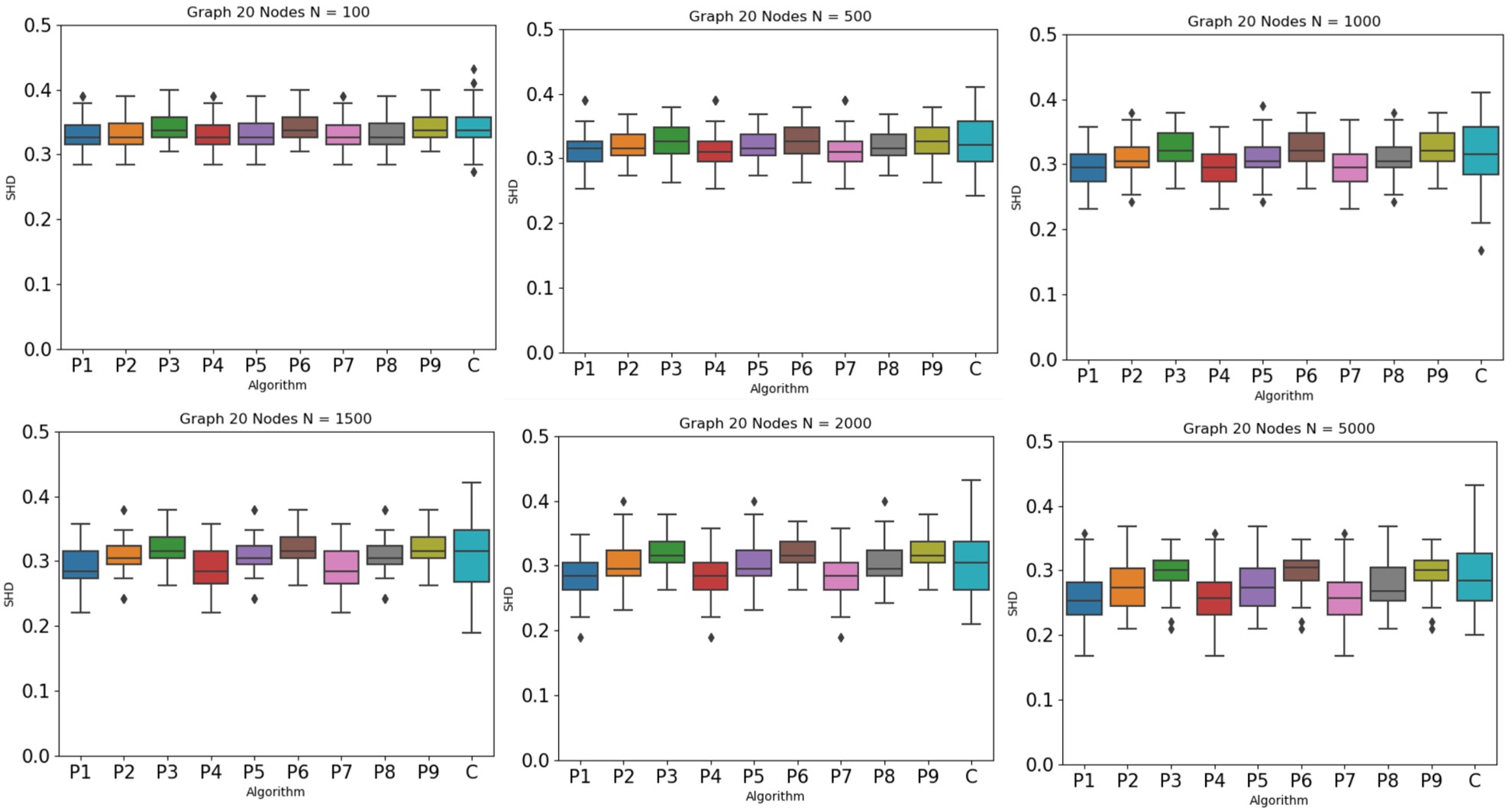}
	\caption{KAPC (P1-P9) and Copula PC (C) Graph Nodes = 20}
	\label{fig:KACopulaPC20_Exp1}
\end{figure}

\begin{figure}[ht]
	\centering
	\includegraphics[width=\textwidth]{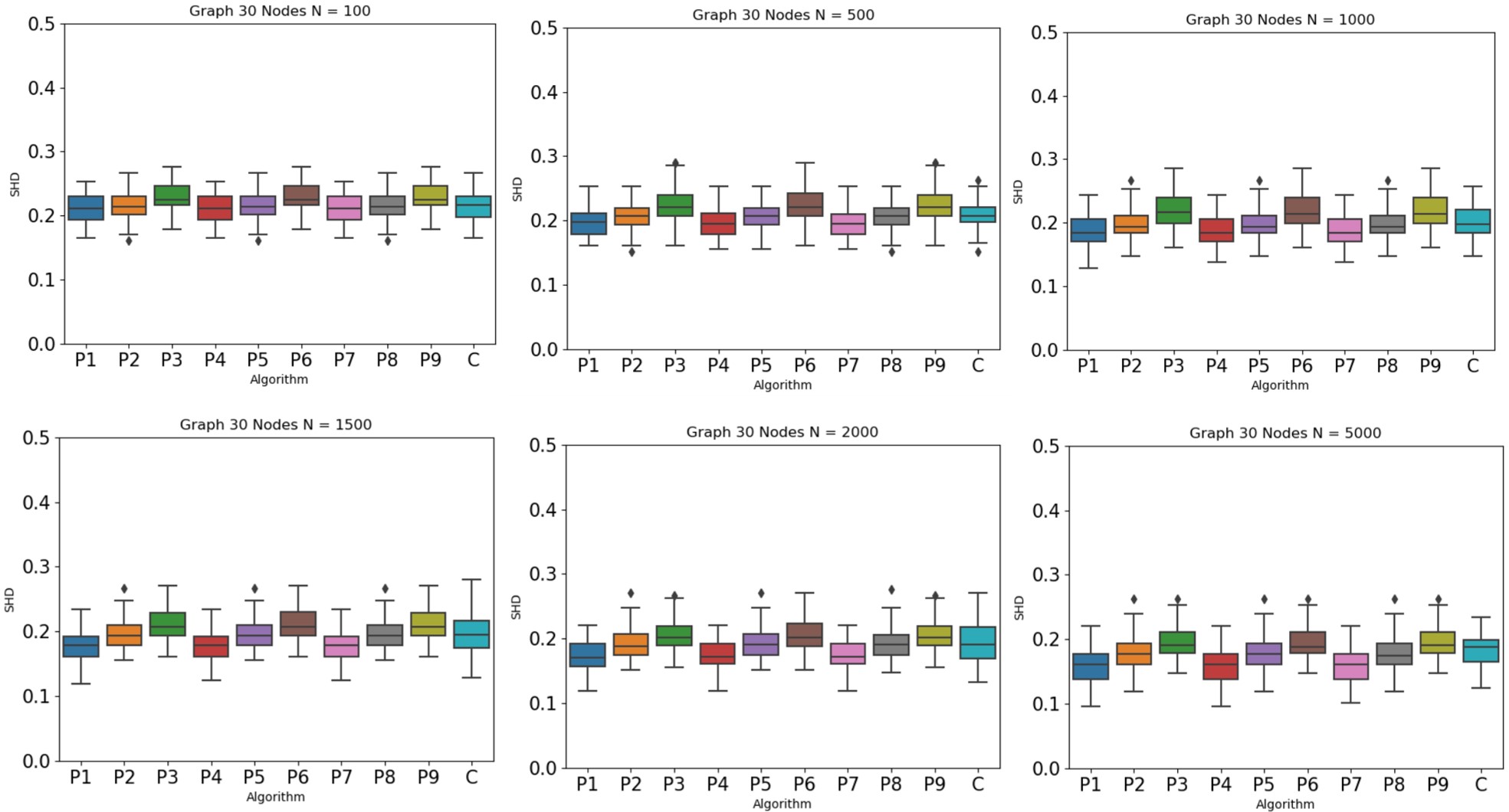}
	\caption{KAPC (P1-P9) and Copula PC (C) Graph Nodes = 30}
	\label{fig:KACopulaPC30_Exp1}
\end{figure}

Figure \ref{fig:KACopulaPC10_Exp1}, \ref{fig:KACopulaPC20_Exp1}, and \ref{fig:KACopulaPC30_Exp1} show SHD score of KAPC and Copula PC from graph containing 10 nodes, 20 nodes, and 30 nodes, respectively. P1-P9 refers to KAPC using particular kernel parameter in Table \ref{tb:widthkern} and C is Copula PC algorithm. In graph 10 nodes, Copula PC algorithm produces lower SHD score than KAPC. However, KAPC generates lower SHD score than Copula PC algorithm in graph 20 nodes and 30 nodes.

\subsection{Experimental Result using Mixed Data: Categorical, Binary, Ordinal, and Continuous Variables}

We only run KAPC and KAFCI to learn causal graphs from mixed datasets containing categorical, binary, ordinal and continuous because Copula PC algorithm implements rank method that is not suitable for categorical variable. In the previous research, it was explained that the Copula PC algorithm can be used to learn causal graph from binary, ordinal and continuous variables \citep{Cui16}. 

In the second experiment, kernel matrices for binary and categorical variables are computed using Categorical kernel. Meanwhile, kernel matrices for ordinal and continuous variables are computed using RBF kernel. The kernel alignment is computed from set of kernel matrices, then it is used as an input for conditional independence test for KAPC. 

In the third experiment, some variables are deleted randomly from the datasets that represent latent variables. Kernel matrices and kernel alignment for conditional independence test for FCI are computed from the remaining variables. The true PAGs are generated from true graphs after removing some variables. The SHD score is computed by comparing the true PAG and the learned PAG.

\begin{figure}[ht]
	\centering
	\includegraphics[width=\textwidth]{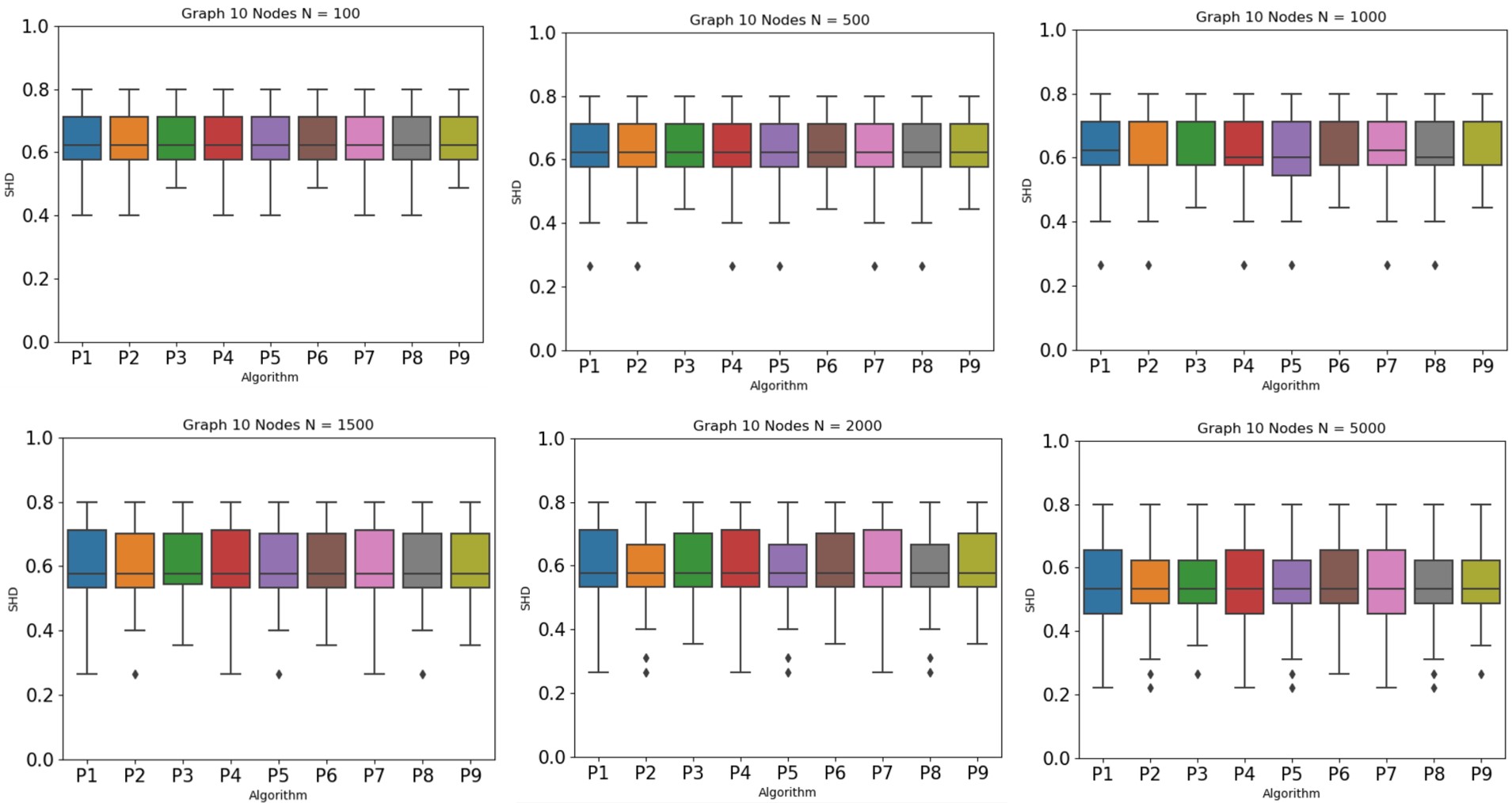}
	\caption{KAPC Graph Nodes = 10}
	\label{fig:KAPC10_Exp3}
\end{figure}

\begin{figure}[ht]
	\centering
	\includegraphics[width=\textwidth]{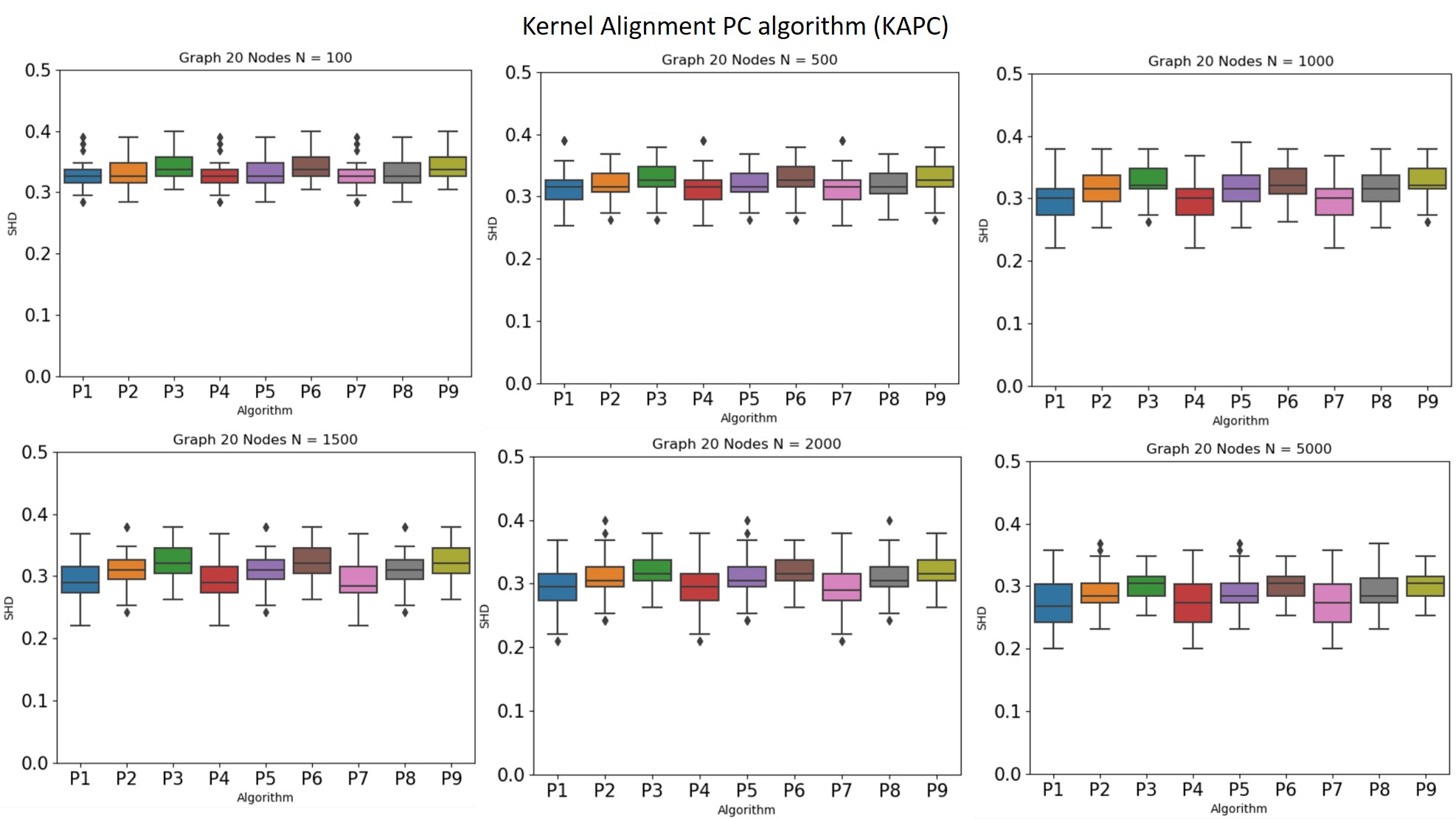}
	\caption{KAPC Graph Nodes = 20}
	\label{fig:KAPC20_Exp3}
\end{figure}

\begin{figure}[ht]
	\centering
	\includegraphics[width=\textwidth]{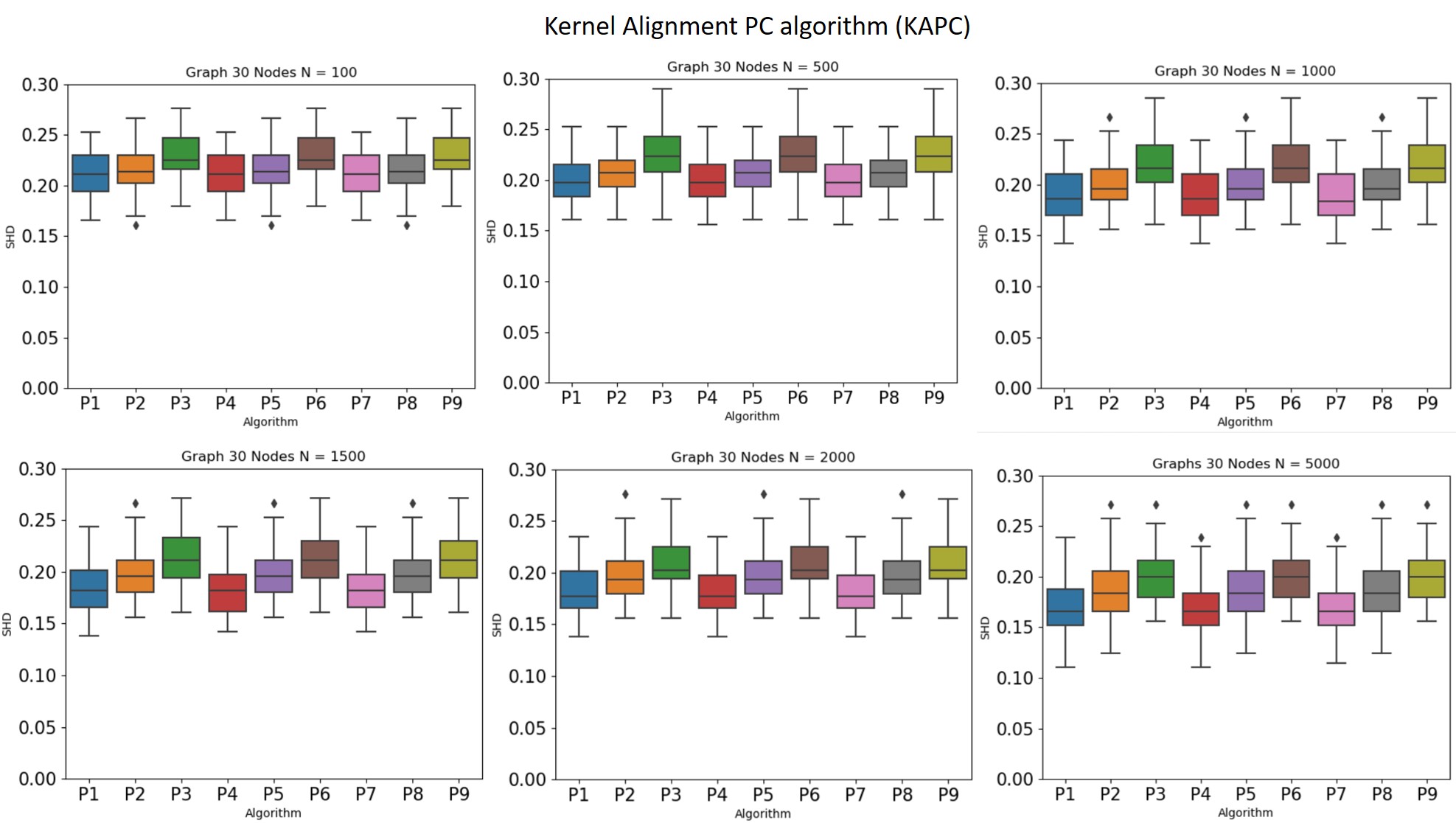}
	\caption{KAPC Graph Nodes = 30}
	\label{fig:KAPC30_Exp3}
\end{figure}

\begin{figure}[ht]
	\centering
	\includegraphics[width=\textwidth]{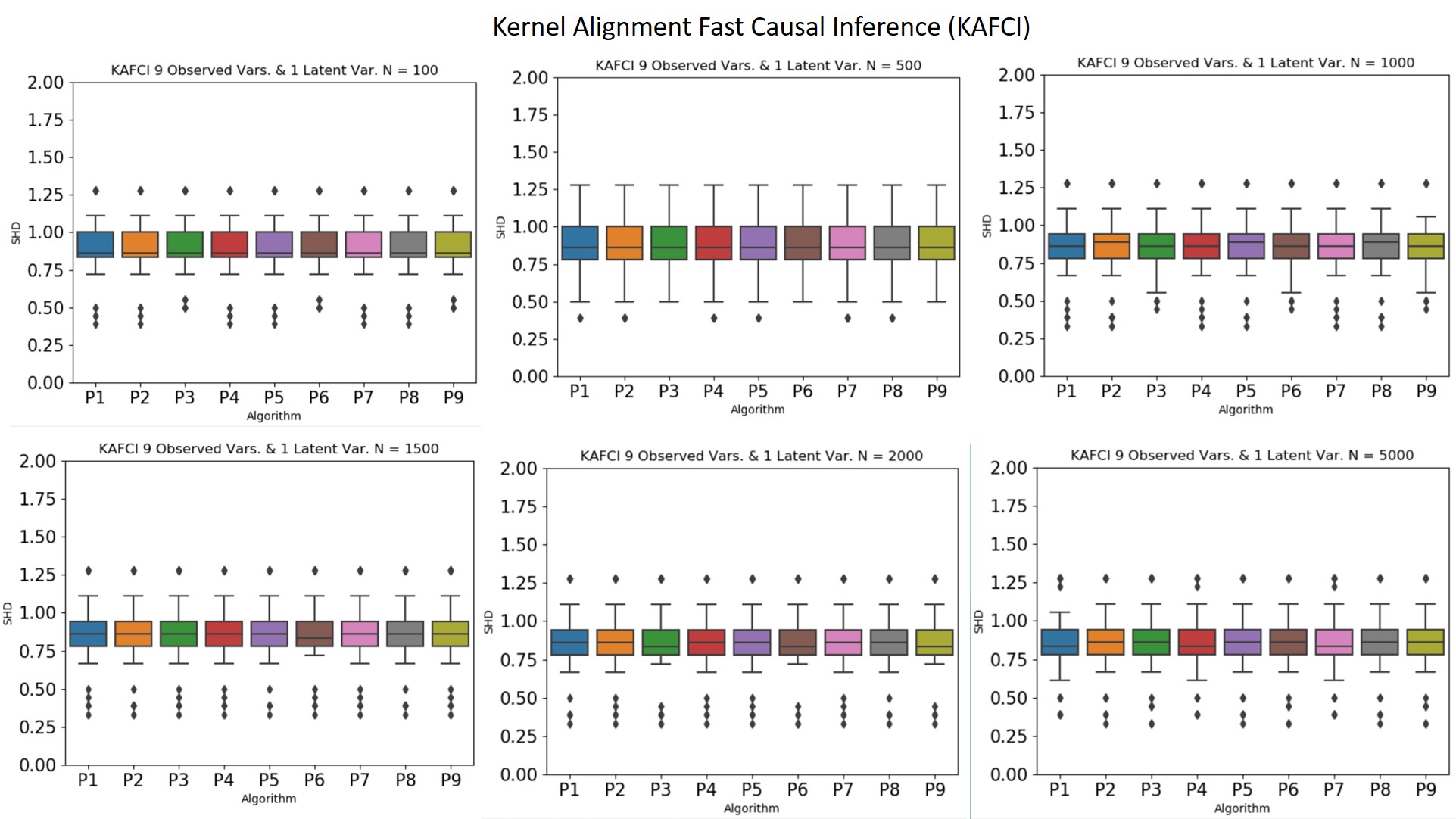}
	\caption{KAFCI Graph 9 Observed Vars. \& 1 Latent Var.}
	\label{fig:KAFCI10_Exp4}
\end{figure}

\begin{figure}[ht]
	\centering
	\includegraphics[width=\textwidth]{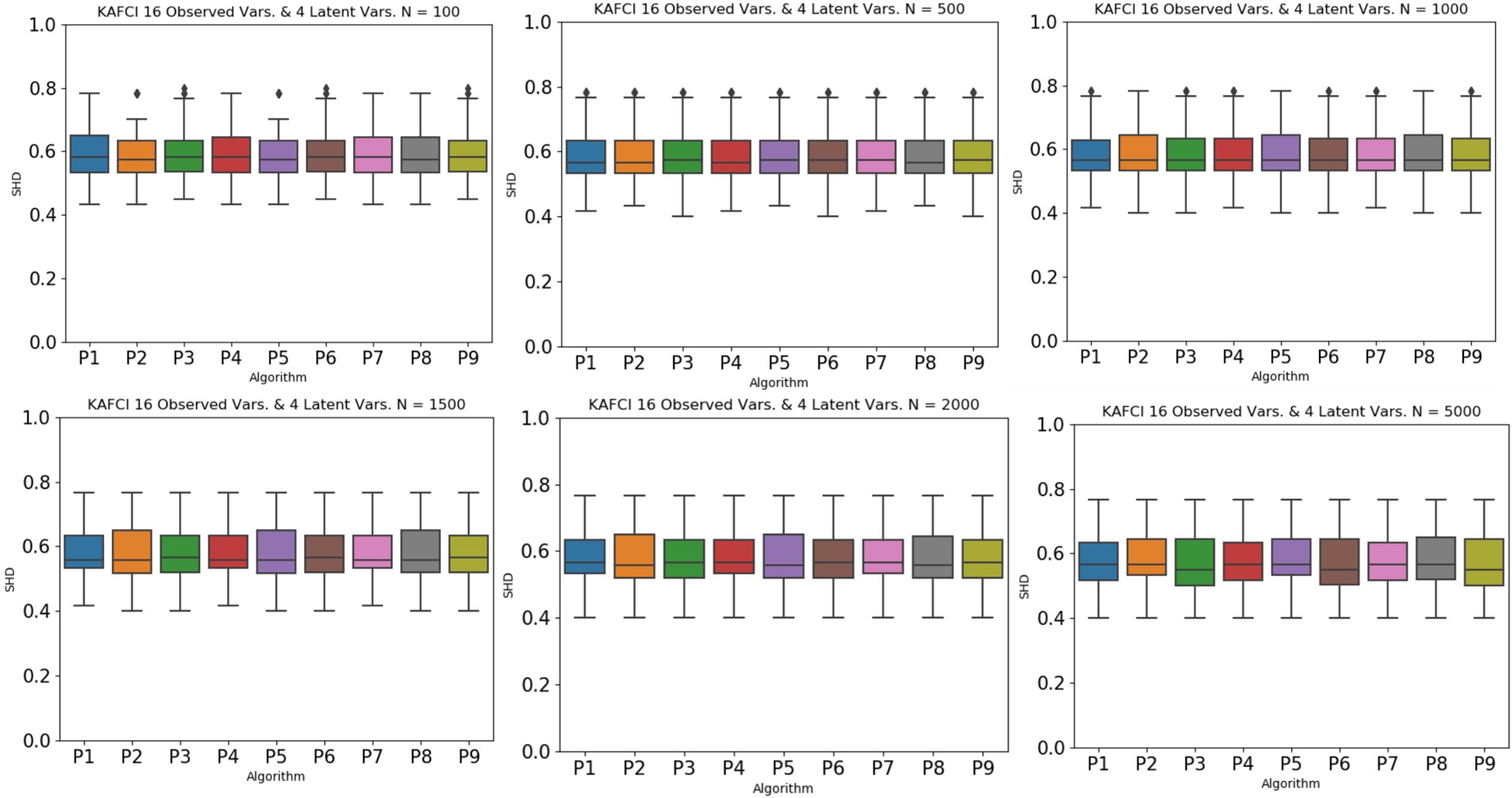}
	\caption{KAFCI Graph 16 Observed Vars. \& 4 Latent Vars.}
	\label{fig:KAFCI20_Exp4}
\end{figure}

\begin{figure}[ht]
	\centering
	\includegraphics[width=\textwidth]{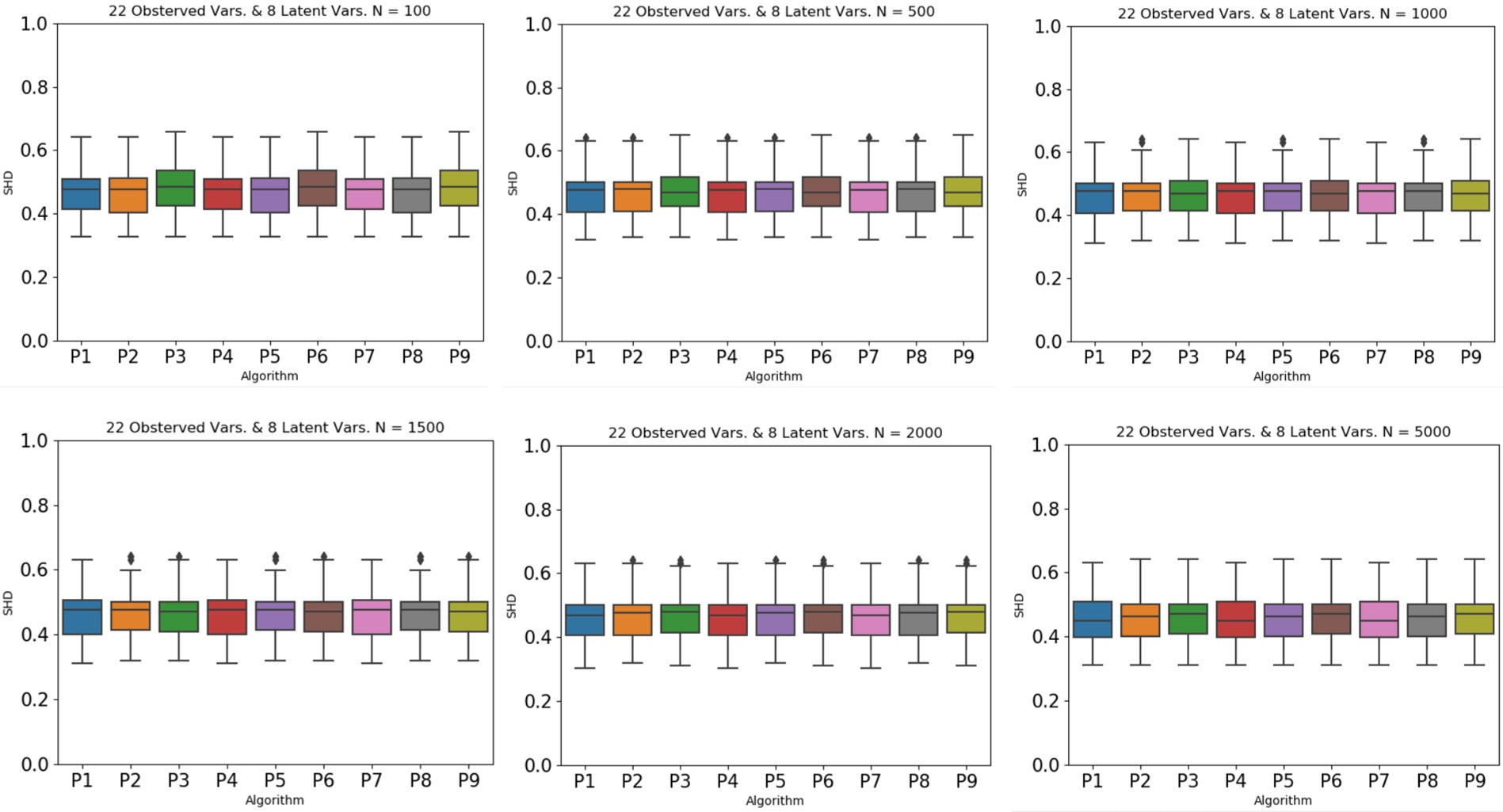}
	\caption{KAFCI Graph 22 Observed Vars. \& 8 Latent Vars.}
	\label{fig:KAFCI30_Exp4}
\end{figure}

Figure \ref{fig:KAPC10_Exp3}, \ref{fig:KAPC20_Exp3}, and \ref{fig:KAPC30_Exp3} show SHD score of KAPC for graphs 10, 20, and 30 nodes, respectively. KAPC produces a lower SHD score when it is given parameter value P1, P4 and P7.  Figure \ref{fig:KAFCI10_Exp4}, \ref{fig:KAFCI20_Exp4}, and \ref{fig:KAFCI30_Exp4} show SHD score of KAFCI for graphs 10, 20, and 30 nodes, respectively. The results show that different values of kernel parameter do not produce different results. 

\begin{figure}[ht]
	\centering
	\includegraphics[width=\textwidth]{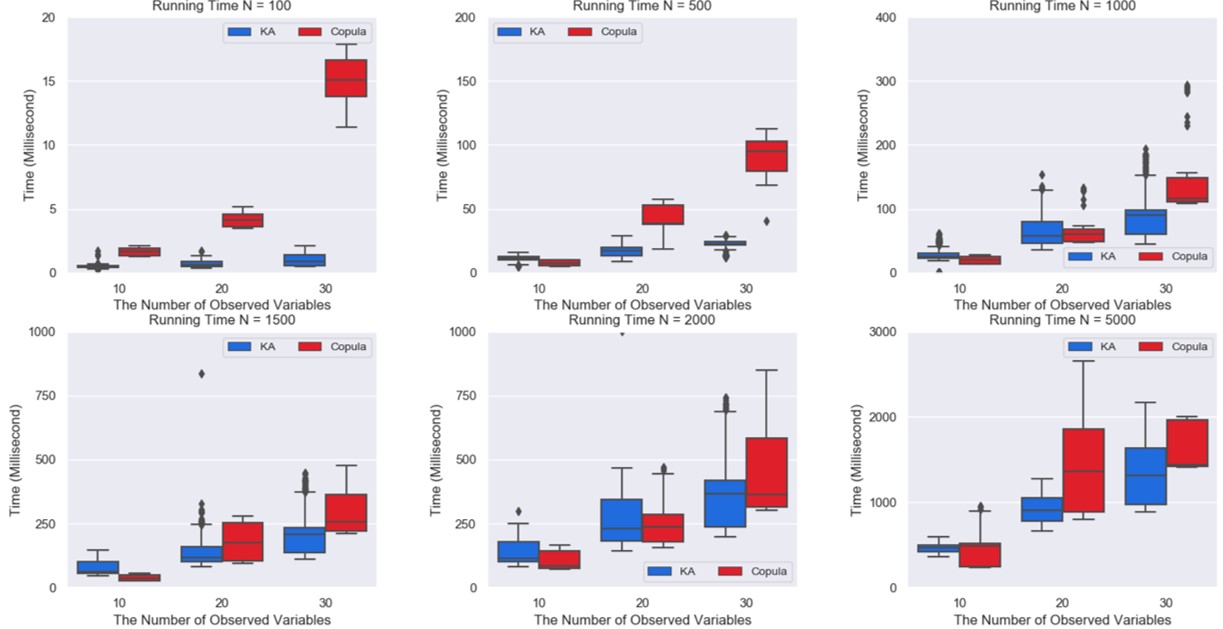}
	\caption{Running Time}
	\label{fig:RunningTime}
\end{figure}

Suppose, the pseudo-correlation matrix is a term that is used for correlation matrix that is computed from a kernel-based approach and the Copula model.  Both kernel-based approach and Copula model compute a pseudo-correlation matrix and use it as an input for a conditional independence test in PC algorithm. The running time is the only time to compute a pseudo-correlation matrix without including time for generating the graph. Figure \ref{fig:RunningTime} shows comparison running time of kernel-based approach and copula approach. 

\section{Discussion}

The main reason to develop kernel-based approach provides the procedure to treat the categorical variable commensurates the binary, ordinal and continuous variable in inferring a causal graph from a mixed data using the existing applications of PC algorithm and FCI. Conditional independence test for PC algorithm and FCI in this research need correlation matrix and the number of data points. Kernel Alignment is applied to compute a pseudo-correlation matrix then it is participated for conditional independence test. Kernel-based approach uses $n$ data points for conditional independence test, where $n$ is the number of data points in dataset. The Copula PC algorithm computes degree of freedom and uses it as the number of data points for conditional independence test \citep{Cui16}. It uses the number of data points around 80\% of the actual number of data points. PC Algorithm and FCI work well if they are given sufficient data. It is believed that the more data points support the algorithms produce more accurate learned graphs. The experimental results support this claim because the more data points used to learn the graph, the learned graphs have smaller SHD score. The smaller SHD score means that the learned graph has less mismatch to the true graph. 

According to the experiment part 1, the experiment results show that Copula PC algorithm works better than KAPC for graph 10 nodes. However, KAPC has a slightly better performance than Copula PC for graph 20 and 30 nodes. It is represented by SHD score KAPC is lower than SHD score Copula PC algorithm. The advantage of a KAPC is this proposed method works for categorical variables as well as binary, ordinal, and continuous variables, so it is a step further than the Copula PC algorithm. Generally, the kernel-based approach is running faster than the Copula model. Furthermore, Copula PC algorithm is not working for 48 datasets due to failure to generate the scale matrices. The scale of matrix in Copula PC algorithm acts as a correlation matrix for a conditional independence test.  There is no failure in computing the pseudo-correlation matrix using the proposed method during the experiment. We do not observe more about the reason for failure because it is not our main concern. 

The issue in applying the kernel-based approach is how to choose the best value of kernel parameter for the kernel function. The kernel-based approach can be implemented to learn causal graph from mixed data as long as there is a suitable kernel function for particular variables. The experiments using 9 different values of parameters show that KAPC produces graphs with lower SHD score when it is given proper kernel parameter values. The experiments show that parameter P1, P4 and P9 are successfully produces learned graphs with lower SHD score. Three of them use RBF kernel $\sigma = 0.001$. 

The experiment results using KAFCI show that the SHD score of 9 different kernel parameter are similar. The SHD score of KAFCI is higher than SHD score KAPC. The learned graph of KAFCI is represented by a PAG which have six different types of edges. A learned graph represented by PAG has a high mismatch to its PAG because of the same edge might have different marks. In an example, the true edge is $A \leftrightarrow B$ but the learned edge is $A \circ\!\!\!\rightarrow B$, then the SHD identifies they are a mismatch.  

The kernel-based approach gives a simple solution for learning causal graphs using the existing application of the PC algorithm and FCI when the dataset containing categorical, binary, ordinal, and continuous variables. Moreover, the kernel-based approach is possible to use for other data types throughout there is a suitable kernel function for these variables. This proposed method works well for the datasets without missing values. 

\section{Conclusion}

In conclusion, kernel-based approach is a promising method for inferring a causal graph from mixed data containing categorical, binary, ordinal, and continuous variables using PC algorithm and FCI. Kernel-based approach offers better treatment for the categorical variable in the mixed data which cannot be handled properly using the Copula model developed by \citet{Cui16}. It also runs faster than Copula model introduced by \citet{Cui16}. 

For the future research, it needs to explore how to automatically select the proper value of kernel width for RBF kernel and Categorical kernel. Furthermore, it requires to investigate how to choose the most suitable kernel function for each variable. In fact, the dataset might contain missing value data and the current method is developed for the dataset without missing values. It is necessary to improve the performance of the kernel-based approach, so it can be used to learn a causal graph from a mixed data contains missing values.




\newpage



%




\clearpage
\bibliographystyle{apalike}
\bibliography{Bibliography_KA}

\end{document}